\documentclass{article}

 \PassOptionsToPackage{numbers,compress,sort}{natbib}

\usepackage[preprint]{nips_2018}

\usepackage[utf8]{inputenc} 
\usepackage[T1]{fontenc}    
\usepackage{hyperref}       
\usepackage{url}            
\usepackage{booktabs}       
\usepackage{amsfonts}       
\usepackage{nicefrac}       
\usepackage{microtype}      

\usepackage{amsmath}

\DeclareMathOperator*{\argminA}{arg\,min}
\usepackage{amssymb} 
\usepackage{subcaption}
 \usepackage[pdftex]{graphicx}

\title{A Neural Network Based Method to Solve Boundary Value Problems}

\author{
  Sethu Hareesh Kolluru\\
  \texttt{hareesh@stanford.edu} \\
}

\begin{document}

\maketitle

\begin{abstract}
A Neural Network (NN) based numerical method is formulated and implemented for solving Boundary Value Problems (BVPs) and numerical results are presented to validate this method by solving Laplace equation with Dirichlet boundary condition and Poisson's equation with mixed boundary conditions. The principal advantage of NN based numerical method is the discrete data points where the field is computed, can be unstructured and do not suffer from issues of meshing like traditional numerical methods such as Finite Difference Time Domain or Finite Element Method. Numerical investigations are carried out for both uniform and non-uniform training grid distributions to understand the efficacy and limitations of this method and to provide qualitative understanding of various parameters involved. 
\end{abstract}

\section{Background and Motivation}
    
In the field of ElectroMagnetics (EM), Boundary Value Problems (BVPs) are the problems for which the EM field in a given region of space is determined from a knowledge of the field over the boundary of the region \citep{BVP}. Traditional numerical methods such as Finite Difference Time Domain (FDTD) and Finite Element Methods (FEM) are typically employed to solve BVPs. However, these methods involve discretization of the domain to reduce it to higher-order system of linear algebraic equations and solving for them. As such, these methods are not local i.e. they do not give the value of the solution directly at an arbitrary point, where the field needs to be determined, but its value should be extracted from the complete field solution and hence are not amenable to parallel processing. 
    
    Neural Network (NN) based numerical method provides an alternate approach to solving BVPs \citep{Lagaris},\citep{Mcfall}. The principal advantages of the NN based numerical method are the discrete data points where field is computed, can be unstructured and therefore, the issues of meshing (uniform/non-uniform) are not a factor; the solutions are in a differentiable, closed analytic form which avoids the need to interpolate between data points where solutions are obtained using other methods. Also, recent advances in the field of Machine Learning (ML) and Artificial Intelligence (AI) has jump started the design and implementation of computer architectures that are optimized to implement training and inference tasks more efficiently. Since, this NN based numerical method is inherently parallel and hence can be efficiently implemented on parallel architectures, 
 it stands to gain from advances in AI based computer architectures.
	In this study, NN based field computation is formulated and presented for BVPs with Dirichlet boundary condition and BVPs with mixed boundary conditions on a uniform rectangular grid. Multiple non-uniform training grid distributions are also explored in this study  to showcase that this method is not limited by the domain discretization like traditional methods.


\section{Problem Statement}
NN based method is developed, implemented and investigated for solving Laplace equation with Dirichlet boundary condition and Poisson's equation with mixed boundary condition. Numerical investigations are carried out to
understand efficacy 
of this method and to provide qualitative understanding
of various parameters involved.

\section{Formulation}
BVPs, where the EM field $\psi(x)$ is given by the linear second order partial differential equation (PDE), are considered in this study \cite{Lagaris}
\begin{eqnarray}
G(x,\psi(x),\nabla\psi(x),\nabla^2\psi(x))&=& 0,\quad \forall x\in D
\label{l2pde}
\end{eqnarray}
subject to boundary condition (B.Cs). Here $x =(x_1,x_2,\cdots,x_n) \in\mathbb{R}^n$ and $D\subset{\mathbb{R}^n}$. To solve this BVP using NN based method, a trial form of the solution is constructed, which is written as sum of two parts: the first part satisfies the boundary conditions and contains no adjustable parameters and the second part which involves a feed forward neural network with adjustable parameters is constructed so as not to contribute to boundary conditions.
\begin{eqnarray}
\psi_t(x,W,b)&=&\hat{\psi}(x)+F(x)N(x,W,b)
\end{eqnarray}
$N(x,W,b)$ is a feed forward NN with weights $W$ and biases $b$. $F(x)$ is chosen such that second part does not contribute to boundary conditions.

In such a case, the task of field computation reduces to learning the NN parameters, $W$ and $b$, which is done by first transforming the equation (\ref{l2pde}) to a discretized version and applying it at discretized domain, $\hat{D}=\{x^{(i)}\in D; i=1,\cdots,m\}$.
\begin{eqnarray}
G(x^{(i)},\psi(x^{(i)}),\nabla\psi(x^{(i)}),\nabla^2\psi(x^{(i)}))&=& 0,\quad \forall x^{(i)}\in D \nonumber 
\end{eqnarray}
and then training the NN, where the PDE error or cost corresponding to $x^{(i)}$
 has to become zero. 
\small
\begin{eqnarray}\small
W,b=\argminA_{W,b}{} G(x^{(i)},\psi_t(x^{(i)},W,b),\nabla\psi_t(x^{(i)},W,b),\nabla^2\psi_t(x^{(i)},W,b))^2 \nonumber
\end{eqnarray}
\normalsize
Note that the computation of this cost function involves not only the network output, but also the derivatives of the output with respect to any of its inputs. 

\subsection{Neural Network Architecture}
\begin{figure}[!hbt]
		\centering
		\includegraphics[width=.7\linewidth]{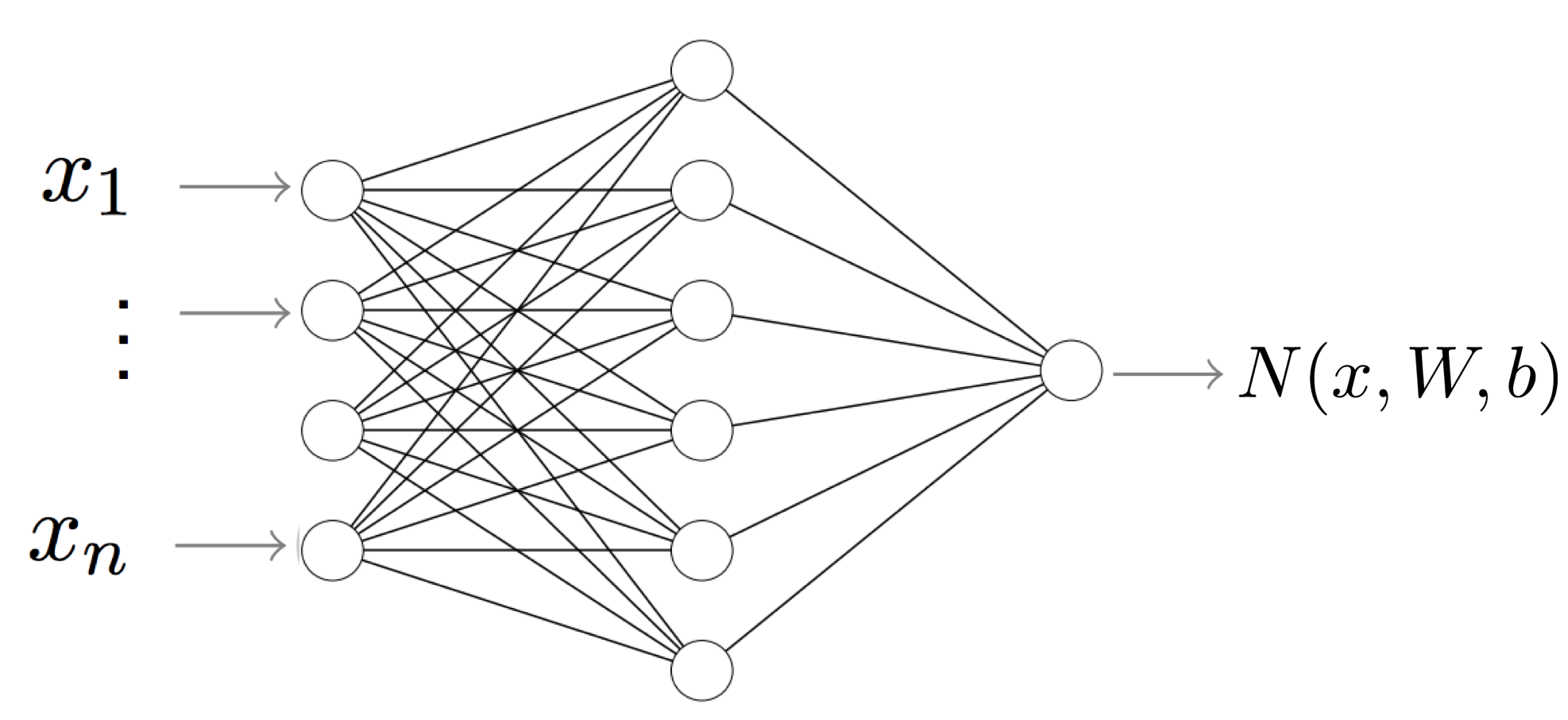}
		\caption{Neural Network Architecture with $n$ input nodes, $H$ hidden nodes and $1$ output node.}
		\label{fig:NN}

	\end{figure}

The neural network that will be implemented for this solution is a $3$-layer network with an input layer ($n+1$ nodes), a hidden layer ($H$ nodes) and an output layer ($1$ node) as shown in Fig.\ref{fig:NN}. Sigmoid function $(\sigma(.))$ will be used as activation function for the hidden layer.
\begin{eqnarray}
h&=&W^{[1]}x+b^{[1]} \nonumber \\
N&=&W^{[2]}\sigma(h) \nonumber
\end{eqnarray}
where $W^{[1]} \in \mathbb{R}^{H\times n}, W^{[2]} \in \mathbb{R}^{1\times H}$ and $h,b^{[1]} \in \mathbb{R}^{H\times 1}$. $W^{[1]}_j$ is used to denote the $j^{th}$ column of $W^{[1]}$.

\subsection{Network Output Derivative Computation}
For this neural network, it can be shown that 
\begin{eqnarray}
\frac{\partial^{\lambda_1}}{\partial x_1^{\lambda_1}}\frac{\partial^{\lambda_2}}{\partial x_2^{\lambda_2}}\cdots \frac{\partial^{\lambda_n} }{\partial x_n^{\lambda_n}} N&=&\sum_{i=1}^H W^{[2]}_i\left(\prod_{j=1}^n(W^{[1]}_{ij})^{\lambda_j}\right)\sigma^{(\lambda)}(h_i) \nonumber
\end{eqnarray}
where $\lambda=\sum_{i=1}^n\lambda_i$ and $\sigma^{(\lambda)}(h_i)$ denotes the $\lambda$th order derivative of the sigmoid.

Therefore, first order derivate of $N$, with respect to any input parameter $x_j$ is given by 
\begin{eqnarray}
\frac{\partial N }{\partial x_j} &=& \sum_{i=1}^H W^{[2]}_i(W^{[1]}_{ij})\sigma^{(1)}(h_i) = \left(W^{[2]}\circ W^{[1]}_j\right)) \sigma^{(1)}(h_i) \nonumber
\end{eqnarray}
which can be interpreted as the output of the feedforward neural network of same schematic, where the activation function for the hidden layer is given by the first order derivative of sigmoid, instead of sigmoid and the $W^{[2]}$ replaced by $W^{[2]}\circ W^{[1]}_j$. 

Similarly, the second order derivative of $N$ with respect to $x_j$ can be interpreted as the output of a feedforward neural network with the same architecture, where the activation function for the hidden layer is given by the second order derivative of sigmoid and the $W^{[2]}$ replaced by $W^{[2]}\circ (W^{[1]}_j \circ W^{[1]}_j)$. 
\begin{eqnarray}
\frac{\partial^2 N }{\partial x_j^2} &=& \sum_{i=1}^H W^{[2]}_i(W^{[1]}_{ij})^2\sigma^{(2)}(h_i) \nonumber \\&=& \left(W^{[2]}\circ W^{[1]}_j\circ W^{[1]}_j\right)) \sigma^{(2)}(h_i) \nonumber
\end{eqnarray}
For the cost function, which includes network output as well as derivatives of network output, these interpretations become extremely useful during training, when cost function is being optimized. 

\section{Implementation}
The formulated method is implemented to find solution of two BVPs - Laplace equation with Dirichlet boundary condition and Poisson's equation with mixed boundary condition. In both examples, the domain is chosen to be a square $D=[0,1]\times [0,1]$. 
Neural Network has been implemented using Tensor Flow framework and optimized using Stochastic Gradient Descent (SGD) with annealing learning rate and regularization.

\subsection{Laplace Equation with Dirichlet boundary condition}
Electrostatic potential distribution inside a rectangular region where the potential on the boundary is specified, is given by the Laplace equation with Dirichlet boundary condition in a $2D$ rectangular region \cite{janaswamy}. NN based method is used to compute the solution and compared with the analytical solution \cite{stanford}.
\begin{eqnarray}
\nabla^{2}\psi(x)=0, \quad \forall x \in D
\end{eqnarray}
The boundary conditions are
\begin{eqnarray}
\psi(x)&=&0, \quad \forall x \in \{(x_1,x_2)\} \in \begin{cases}
 \partial D| x_1=0\\
 \partial D| x_1=1\\
 \partial D| x_2=0
\end{cases} \nonumber \\
\psi(x)&=&sin(\pi x_1),\quad \forall x \in \{(x_1,x_2)\in\partial D|x_2=1 \} \nonumber
\end{eqnarray}

\begin{figure}[h]
\centering
\begin{minipage}{.5\textwidth}
  \centering
  \includegraphics[width=1.1\linewidth]{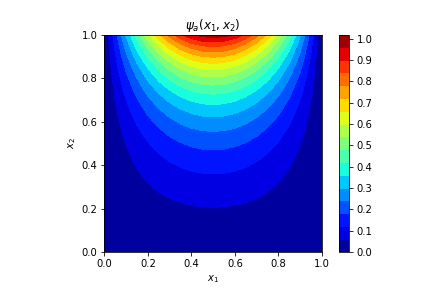}
  \captionof{figure}{Analytical solution}
  \label{fig:laplace_a}
\end{minipage}%
\begin{minipage}{.5\textwidth}
  \centering
  \includegraphics[width=1.1\linewidth]{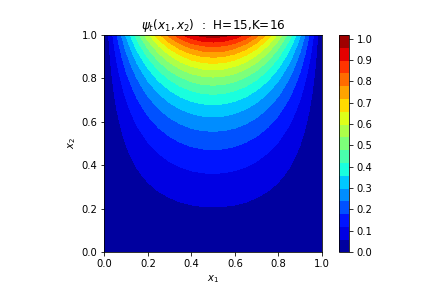}
  \captionof{figure}{NN based solution}
  \label{fig:laplace_nn}
\end{minipage}
\end{figure}

\begin{figure}[!hbt]
		\begin{center}
		\includegraphics[width=.56\linewidth]{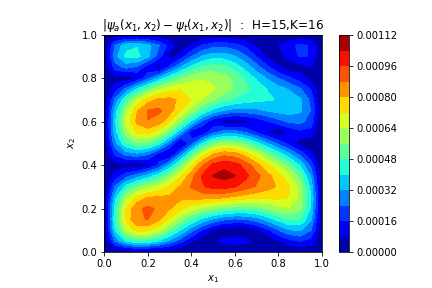}
		\caption{$E_{abs}$, Laplace equation}
		\label{fig:delta_1}
		\end{center}
	\end{figure}

The analytical solution is 
\begin{eqnarray}
\psi_a(x)&=&\frac{1}{e^\pi-e^{-\pi}}sin(\pi x_1)(e^{\pi x_2}-e^{-\pi x_2})
\end{eqnarray}
The trial solution constructed for NN based method is 
\begin{eqnarray}
\psi_t(x)&=&x_2sin(\pi x_1)+x_1(1-x_1)x_2(1-x_2)N(x,W,b) \nonumber 
\end{eqnarray}
In this case, the cost function is given by
\begin{eqnarray}
&&{-\pi^2 x_2 sin(\pi x_1)+} \nonumber \\
&&{x_2(1-x_2)\left(x_1(1-x_1)\frac{\partial^2 N}{\partial x_1^2} +(2-4x_1)\frac{\partial N}{\partial x_1}-2N\right)+} \nonumber\\
&&{x_1(1-x_1)\left(x_2(1-x_2)\frac{\partial^2 N}{\partial x_2^2} +(2-4x_2)\frac{\partial N}{\partial x_2}-2N\right)}\nonumber 
\end{eqnarray}
The analytical solution (Fig. \ref{fig:laplace_a}) and the NN based solution (Fig. \ref{fig:laplace_nn}) computed by minimizing the cost function when $K=16$ and $H=15$ are shown below
The absolute value of the delta between the two solutions, $|\psi_a(x)-\psi_t(x)|$ is also plotted in Fig. \ref{fig:delta_1}.

\subsection{Poisson's equation with mixed boundary condition}
\begin{figure}[!hbt]
		\begin{center}
		\includegraphics[width=.56\linewidth]{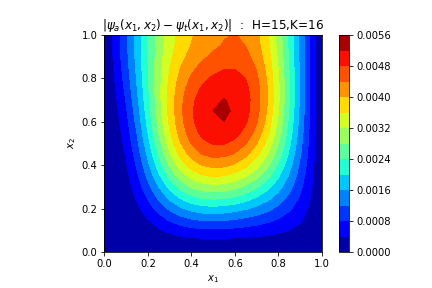}
		\caption{$E_{abs}$, Poisson's equation}
		\label{fig:delta}
		\end{center}
\end{figure}

Electrostatic potential in the presence of charge distribution inside a rectangular region where the potential is specified on a section of the boundary and the gradient of the potential is specified on the rest is considered here. 
\begin{eqnarray}
\nabla^{2}\psi(x)=(2-\pi^2x_2^2)sin(\pi x_1), \quad \forall x \in D
\end{eqnarray}
The mixed boundary conditions are
\begin{eqnarray}
\psi(x)&=&0, \quad \forall x \in \{(x_1,x_2)\in
\begin{cases}
 \partial D| x_1=0\\
 \partial D| x_1=1\\
 \partial D| x_2=0
\end{cases} \nonumber \\
\frac{\partial \psi(x)}{\partial x_2}&=&2sin(\pi x_1),\quad \forall x \in \{(x_1,x_2)\in\partial D_n|x_2=1 \} \nonumber
\end{eqnarray}
The analytical solution is 
\begin{eqnarray}
\psi_a(x)&=&x_2^2 sin(\pi x_1)
\end{eqnarray}
The trial solution constructed for NN based method is 
\begin{eqnarray}
\psi_t(x)&=&2x_2sin(\pi x_1)\nonumber \\
&&+x_1(1-x_1)x_2\left[N(x_1,x_2,W,b)\right] \nonumber \\
&&-x_1(1-x_1)x_2\left[ N(x_1,1,W,b)+\frac{\partial N(x_1,1,W,b)}{\partial x_2}\right] \nonumber 
\end{eqnarray}
In this case, the cost function is given by
\begin{eqnarray}
&&-(2-\pi^2 x_2)sin(\pi x_1)-2\pi^2 x_2 sin(\pi x_1)+ \nonumber \\
&&x_2x_1(1-x_1)\left[\frac{\partial^2 N(x_1,x_2)}{\partial x_1^2}-\frac{\partial^2 N(x_1,1)}{\partial x_1^2} -\frac{\partial^3 N(x_1,1)}{\partial x_1^2 \partial x_2}\right]+\nonumber\\
&&2x_2(1-2x_1)\left[\frac{\partial N(x_1,x_2)}{\partial x_1}-\frac{\partial N(x_1,1)}{\partial x_1} -\frac{\partial^2 N(x_1,1)}{\partial x_1 \partial x_2}\right]-\nonumber\\
&&2x_2\left[N(x_1,x_2)-N(x_1,1)-\frac{\partial N(x_1,1)}{\partial x_2}\right]+\nonumber \\
&&x_2x_1(1-x_1)\left[\frac{\partial^2 N(x_1,x_2)}{\partial x_2^2}\right] 
+2x_1(1-x_1)\left[\frac{\partial N(x_1,x_2)}{\partial x_1}\right]
\end{eqnarray}
The analytical solution and the NN based solution computed by minimizing the cost function when $K=16$ and $H=15$ are shown below
The absolute value of the delta between the two solutions, $|\psi_a(x)-\psi_t(x)|$, with a maximum value of $0.0056$, plotted in Fig. \ref{fig:delta} showcases good agreement between the two solutions.
\section{Numerical Investigation on Error Properties}
Let $\hat{D}_{train}=\{x^{(i)}_{train}\in D; i=1,\cdots,m_{train}\}$ and $\hat{D}_{test}=\{x^{(i)}_{test}\in D; i=1,\cdots,m_{test}\}$ represent the set of training and testing dataset respectively. For the case of uniform discretization of $2D$ domain with a training grid resolution, $K$, $m_{train}=K^2$ and for the non-uniform discretization as well, the term training grid resolution, $K$, is meant to indicate that $m_{train}=K^2$ 
The testing grid resolution for all the experiments is kept fixed at $21$, therefore the testing dataset is of the size $m_{test}=441$, where the $2D$ domain is discretized uniformly.

The performance metrics that will be used in this study are the absolute value of the delta between analytical solution and NN based solution
\begin{eqnarray}
E_{abs}=|(\psi_a(x)-\psi_t(x))|
\end{eqnarray}
and relative error norm as given by
\begin{eqnarray}
E_{norm}=\frac{\sqrt{\sum_x(\psi_a(x)-\psi_t(x))^2}}{\sqrt{\sum_x(\psi_a(x))^2}}
\end{eqnarray}

$E_{abs}$ gives us information about the spatial distribution of the divergence between analytical and NN based solution, while the $E_{norm}$ paints an aggregate picture over the entire domain.

\begin{figure}[!hbt]
		\begin{center}
		\includegraphics[width=.56\linewidth]{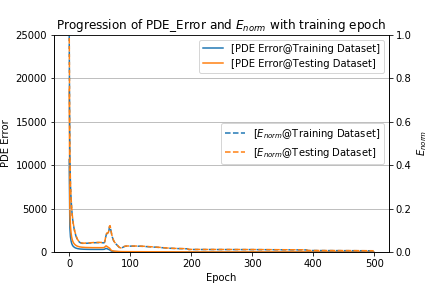}
		\caption{Progression of PDE error and $E_{norm}$ during training}
		\label{fig:convergence}
		\end{center}
	\end{figure}

There is a subtle, but important point that needs to be understood when solving BVPs using NN, $i.e$ training is driven by the cost function (or PDE error), while performance is gauged by the error in the computed field \cite{Mcfall}.

During the optimization of NN, the PDE is more closely satisfied at the training points for each successful training epoch. But the reduction of PDE error at training points has no obvious guarantee that the NN based solution is converging to the analytical solution, even at the training points. In addition to that, NN with poor generalization because of over fitting may not satisfy the PDE in the regions of the domain away from the training points, and hence NN based solution diverges from analytical solution at such points. Convergence theorem outlines that the difference between the analytical and NN based solutions will decrease everywhere in the domain, when the PDE error is also reduced everywhere in the domain \cite{Mcfall2}. In practice, however, PDE error cannot be guaranteed to decrease everywhere in the domain during training. However, convergence plots such as one shown in Fig. \ref{fig:convergence} for the Laplace equation when $H=15,m_{train}=256,m_{test}=441$, should be used to monitor the progression of PDE error and $E_{norm}$ during training to make sure both training and test set error decrease along with PDE error during training. Training grid resolution, Number of hidden nodes, Training dataset distribution and size can be tweaked accordingly, if needed to ensure $E_{norm}$ as well as PDE error decreases during training. 


\begin{figure}[!h]
\centering
\begin{minipage}{.5\textwidth}
  \centering
  \includegraphics[width=\linewidth]{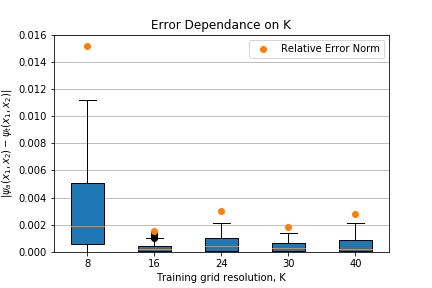}
  \captionof{figure}{Error Dependence on Training Grid \\ Resolution, K}
  \label{fig:Error_K}
\end{minipage}%
\begin{minipage}{.5\textwidth}
  \centering
  \includegraphics[width=\linewidth]{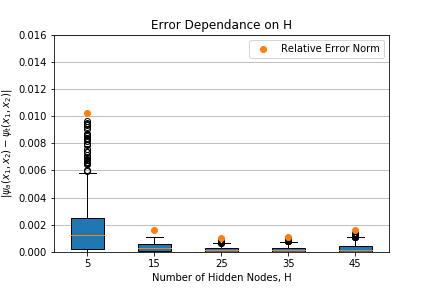}
  \captionof{figure}{Error Dependence on  Number of \\ Hidden Nodes, H}
  \label{fig:Error_H}
\end{minipage}
\end{figure}

%
%

\begin{figure*} [!hbt]
\centering
\begin{subfigure}[t]{0.5\textwidth}
\centering
\includegraphics[width=\textwidth]{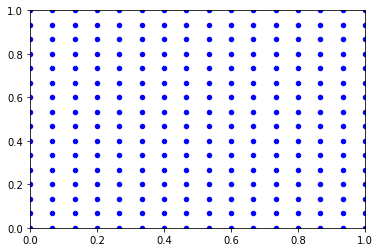}
\caption{Grid 1: Uniform}
\label{fig:grid1}
\end{subfigure}%
%
%
\begin{subfigure}[t]{0.5\textwidth}
\centering
\includegraphics[width=\textwidth]{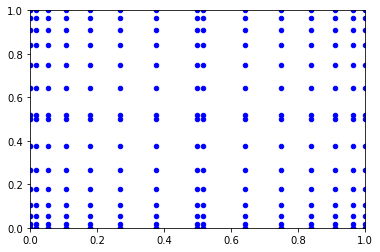}
\caption{Grid 2: Non-Uniform}
\label{fig:grid2}
\end{subfigure}

\bigskip 

\begin{subfigure}[t]{0.5\textwidth}
\centering
\includegraphics[width=\textwidth]{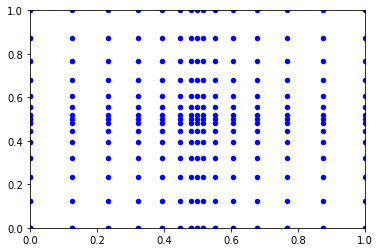}
\caption{Grid 3: Non-Uniform}
\label{fig:grid3}
\end{subfigure}%
%
%
\begin{subfigure}[t]{0.5\textwidth}
\centering
\includegraphics[width=\textwidth]{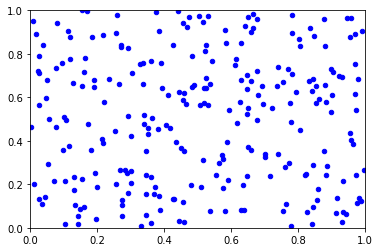}
\caption{Grid 4: Random}
\label{fig:grid4}
\end{subfigure}

\caption{Four different spatial distribution of the training data points investigated}
\label{fig:Grid Experiment}
\end{figure*}

 \begin{figure}[!hbt]
		\begin{center}
		\includegraphics[width=.56\linewidth]{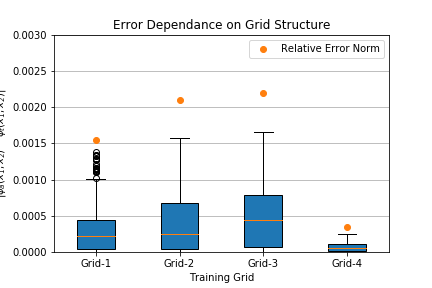}
		\caption{Error dependence on training dataset distribution}
		\label{fig:Error_Grid}
		\end{center}
	\end{figure}
\subsection{Error Dependence on Training Grid Resolution, K}
Neural network was trained for various combinations of the number of hidden nodes, $H$ and training grid resolution, $K$ for the BVP with Laplace equation with Dirichlet boundary condition. For a fixed $H=15$, the variation of test set error when training grid resolution $K={8,16,24,30 \text{ and } 40}$ is shown in Fig.\ref{fig:Error_K}. As $K$ increases from $8$ to $16$, the $E_{abs}$ becomes relatively narrower and $E_{norm}$ decreases from $0.015$ to $0.002$, but plateaus beyond that resolution. Though it is intuitively expected that the test set error improves with increase in the training set size, as the learning model generalizes better, in this particular case, it could also be inferred in the context of the convergence theorem as well. As the PDE error is minimized at more training points, the convergence between analytical and NN based solution gets better. 
\subsection{Error Dependence on Number of Hidden Nodes, H}
To understand the dependence of the test set error on number of hidden nodes, $H$, NN based solution is computed for $H=5,10,15,25,35 \text{ and } 45$, for a fixed $K=16$. Fig.\ref{fig:Error_H} shows how the error improves significantly as $H$ is increased from $5$ to $15$, but stays about the same beyond that. As the computational requirements grow with increase in the number of hidden nodes and/or training grid resolution, and only marginal gains beyond $H=15,K=16$; Hence, this choice of number of hidden nodes and training grid resolution is deemed appropriate for the problem at hand.
\subsection{Error Dependence on Non-Uniform Training Grid}
As mentioned previously, one of the main advantages of NN based method is the training and testing dataset can be totally unstructured and hence they do not suffer from the issue of meshing like other numerical methods. To showcase this, four different grid structures, shown in the Fig.\ref{fig:Grid Experiment} where each dot shows a training point in the training dataset, $m_{train}=256$, are investigated in this study. Fig \ref{fig:grid1} shows ``Grid-1'', the uniform discretization that will be used as baseline in this experiment; Two structured, but different spatial distributions are considered in ``Grid-2" shown Fig \ref{fig:grid2}, which has training points densely distributed closer to the boundary and coarsely away from it and the spatial distribution showing opposite trend in ``Grid-3", shown in Fig \ref{fig:grid3}. Fig \ref{fig:grid4} shows ``Grid-4", the training dataset generated by uniform random distribution in $[0,1]\times[0,1]$.



Fig.\ref{fig:Error_Grid} shows the computed test set error for the four grid structures. ``Grid-4" has the best error performance, followed by ``Grid-1", while ``Grid-2" and "Grid-3" show similar performance. This may be due to fact that the training dataset distribution ``Grid-4" closely resembles the test dataset distribution, compared to the other distributions, as the testing grid is a uniform discretization with a grid resolution of $21$. This demonstrates the NN based method's merits when dealing with unstructured data and its potential to be applied to BVPs with curved or irregular boundaries.


\section{Conclusion and Future work}
In this study, a NN based numerical method has been formulated, implemented and validated for two examples with Dirichlet and mixed boundary conditions. Numerical experiments were carried out to assess the dependence of the error performance on training grid resolution and number of hidden nodes. To showcase the merit of this method, numerical results for structured and unstructured training dataset are presented as well. 

Future efforts could explore application of this method to BVPs with curved or irregular boundaries. Some of the commonly found BVPs in EM are given by PDEs with complex coefficients, which could be the focus of future work as well. 
   


%

\end{document}